# Predicting Cardiovascular Risk Factors from Retinal Fundus Photographs using Deep Learning

Ryan Poplin, MS[1*]
Avinash V. Varadarajan, MS[1*]
Katy Blumer, BS[1]
Yun Liu, PhD[1]
Michael V. McConnell, MD, MSEE[2]
Greg S. Corrado, PhD[1]
Lily Peng, MD, PhD[1**]
Dale R. Webster, PhD[1**]

*Equal contribution
**Equal contribution
[1]Google Research, Google Inc, Mountain View, CA, USA
[2]Verily Life Sciences, South San Francisco, CA, USA; Division of Cardiovascular Medicine, Stanford School of Medicine, Stanford, CA USA

Corresponding Author:
Lily Peng, MD, PhD
Google Research
1600 Amphitheatre Way
Mountain View, CA 94043
lhpeng@google.com

**ABSTRACT**

Traditionally, medical discoveries are made by observing associations and then designing experiments to test these hypotheses. However, observing and quantifying associations in images can be difficult because of the wide variety of features, patterns, colors, values, shapes in real data. In this paper, we use deep learning, a machine learning technique that learns its own features, to discover new knowledge from retinal fundus images. Using models trained on data from 284,335 patients, and validated on two independent datasets of 12,026 and 999 patients, we predict cardiovascular risk factors not previously thought to be present or quantifiable in retinal images, such as such as age (within 3.26 years), gender (0.97 AUC), smoking status (0.71 AUC), HbA1c (within 1.39%), systolic blood pressure (within 11.23mmHg) as well as major adverse cardiac events (0.70 AUC). We further show that our models used distinct aspects of the anatomy to generate each prediction, such as the optic disc or blood vessels, opening avenues of further research.

**INTRODUCTION**

Risk stratification is key to identifying and managing groups at risk for cardiovascular disease, which remains the leading cause of death globally[1]. While the availability of cardiovascular disease risk calculators, such as the Pooled Cohort equations[2], Framingham[3–5] and SCORE[6,7] is fairly widespread, there are many efforts to improve risk predictions. Phenotypic information, particularly of vascular health, may further refine/reclassify risk prediction on an individual basis. Coronary artery calcium[8] is one such example in which additional signals from imaging has been shown to improve risk stratification. Current standard-of-care for screening for cardiovascular disease risk requires a variety of variables derived from patient history and blood samples, such as age, gender, smoking status, blood pressure, body mass index (BMI), glucose, and cholesterol levels[9]. Most cardiovascular risk calculators use some combination of these parameters to identify patients at risk of experiencing either a major cardiovascular event or cardiac-related mortality within a pre-specified time period, such as ten years. However, some of these parameters may be unavailable. For example, in a study from the Practice Innovation and Clinical Excellence (PINNACLE) electronic health record–based cardiovascular registry, the data required to calculate the 10-year risk scores were available for less than 30% of patients[10]. This was largely due to missing cholesterol values[10], which is not surprising given that a fasting blood draw is required to obtain this data. In this situation, BMI can be used in the place of lipids for a preliminary assessment of cardiovascular health[11–13]. Thus, we propose to see if additional signals for risk can be extracted from the retinal images, which can be obtained quickly, cheaply, and noninvasively in an outpatient setting.

Markers of cardiovascular disease, such as hypertensive retinopathy and cholesterol emboli, can often manifest in the eye. Furthermore, because blood vessels can be noninvasively visualized from retinal fundus images, various features in the retina, such as vessel caliber[14–16], bifurcation or tortuosity[17], may reflect the systemic health of the cardiovascular system and future risk. The clinical utility of such features still requires further study. In this work, we demonstrate the extraction and quantification of multiple cardiovascular risk factors from retinal images, using deep learning.

Deep learning is a family of machine learning techniques characterized by multiple computation layers that allow an algorithm to learn the appropriate predictive features based on examples rather than requiring features to be hand-engineered[18]. Recently, deep convolutional neural networks - a special type of deep learning technique optimized for images - have been applied to produce highly accurate algorithms that diagnose diseases from medical images with comparable accuracy to human experts, for example melanoma[19] and diabetic retinopathy (DR)[20].

**Results**

We developed deep learning models using retinal fundus images from 48,101 patients from UK Biobank[21] and 236,234 patients from EyePACS[22], and validated these models using images from 12,026 patients from UK Biobank and 999 patients from EyePACS (Table 1). The mean age was 56.9 ± 8.2 on the UK Biobank clinical validation set dataset and 54.9 ± 10.9 in the EyePACS-2K clinical validation set. The UK Biobank population was predominantly Caucasian while the EyePACS patients were predominantly Hispanic. Hemoglobin A1c measurements were available only in 60% of the EyePACS population. Because this population consisted of mostly diabetic patients presenting for DR screening, the mean HbA1c of this population was 8.2 ± 2.1%, well above the normal range. The UK Biobank population was predominately non-diabetic, and HbA1c levels were not available at the time of this study. Additional patient demographics are summarized in Table 1.

We first tested the ability of our model to predict a variety of cardiovascular risk factors from retinal fundus images (Table 2). Because of the lack of an established baseline for predicting these features from retinal images, we use the average value as the baseline for continuous predictions (e.g., age). The mean absolute error (MAE) for predicting the patient's age was 3.26 years (95% confidence interval (CI): (3.22, 3.31) versus baseline 7.06 (6.98, 7.13) in the UK Biobank validation set. For the EyePACS-2k, the MAE was 3.42 (3.23, 3.61) vs. baseline: 8.48 (8.07, 8.90). The predicted age and actual age have a fairly linear relationship (Figure 1A), which is consistent over both datasets. The algorithm also predicted systolic blood pressure (SBP) better than baseline (Table 2). The predicted SBP increased linearly with actual SBP until approximately 150 mmHg, but leveled off above that value (Figure 1B). We also found that our approach was able to infer ethnicity, another potential CV risk factor[2] (kappa score of 0.60 (95% CI: 0.58-0.63) in the UK Biobank validation set, and 0.75 (0.70-0.79) in the EyePACS-2K validation set).

Given that retinal images alone were sufficient to predict several cardiovascular risk factors to varying degrees, we reasoned that the images could be correlated directly with cardiovascular events. Thus, we next trained a model to predict the onset of major adverse cardiovascular events (MACE) within 5 years. This outcome was available only for one of our datasets, the UK Biobank. Because the UK Biobank is a fairly recent study that recruited relatively healthy participants, MACE were rare (631 events occurred within 5 years of retinal imaging--150 of which were in the clinical validation set). Despite the limited number of events, our model achieved an AUC of 0.70 (95% CI: 0.648, 0.740) from retinal fundus images alone, comparable to the AUC of 0.72 (0.67, 0.76) for the European SCORE risk calculator (Table 3).

Next, we used soft attention (see Methods) to identify the anatomical regions that the algorithm might have been using to make its predictions. A representative example of a single retinal fundus image with accompanying attention maps (also called saliency maps[23]) for each prediction is shown in Figure 2. In addition, for each prediction task, ophthalmologists blinded to the prediction task of the model assessed 100 randomly chosen retinal images to identify patterns in the anatomical locations highlighted by the attention maps for each prediction (Table 4). Encouragingly, the blood vessels were highlighted in the models trained to predict risk factors such as age, smoking, and SBP. Models trained to predict HbA1c tended to highlight the perivascular surroundings, and models trained to predict gender primarily highlighted the optic disc. For other predictions, such as diastolic blood pressure and BMI, the attention masks were non-specific, such as uniform "attention" or highlighting the circular border of the image, suggesting that the signals for those predictions may be distributed more diffusely throughout the image.

**Discussion**

Our results indicate that deep learning of retinal fundus images alone can predict multiple cardiovascular risk factors, including as age, gender, and systolic blood pressure. That these risk factors are core components used in multiple cardiovascular risk calculators indicates that our model can potentially predict cardiovascular risk directly. This is supported by our preliminary results for prediction of MACE.

Encouragingly, the corresponding attention maps also indicate that the neural network model is paying attention to the vascular regions in the retina to predict several variables associated with cardiovascular risk. These attention data, together with the fact that our results are consistent in two separate validation datasets, suggest that the predictions are likely to generalize to other datasets, and indicate pathological phenomena that can be studied further.

Despite these promising results, our study contains several limitations. The dataset size is relatively small for deep learning. In particular, although the AUC for cardiovascular events was comparable to SCORE, the confidence intervals for both methods were wide. A significantly larger dataset or a population with more cardiovascular events may enable more accurate deep learning models to be trained and evaluated with high confidence. In addition, some of the risk factors were not available for both datasets (Table 2). Additional validation of our models on other datasets would be beneficial for these predictions. Furthermore, training with larger datasets and more clinical validation will help determine whether retinal fundus images may be able to augment or replace some of the other markers, such as lipid panels, to yield more accurate score.

To conclude, our study provides evidence that deep learning may uncover additional novel signals in retinal images that will allow for better cardiovascular risk stratification, and suggests avenues of future research into the source of these associations and whether they can be used to better understand and prevent cardiovascular disease.

**METHODS**

**Study participants**

We used two datasets in this study. The first, UK Biobank, is an observational study that recruited 500,000 participants, aged 40-69, across the UK between 2006 and 2010. Each participant was consented and went through a series of health measurements and questionnaires. Each participant also provided blood, urine and saliva samples[21]. However, glucose, cholesterol and hemoglobin A1c (HbA1c) measurements were not available at the time of this study. A total of 67,725 patients[24] also subsequently underwent paired retinal fundus and OCT imaging using a Topcon 3D OCT 1000 Mk2 (Topcon Corporation, Tokyo, Japan). Participants were then followed for health outcomes such as hospitalizations, mortality, and cause of death. The study was reviewed and approved by the North West Multi-Centre Research Ethics Committee. We divided this dataset into a development set to develop our models (80%) and validation set to assess our model's performance (20%).

The second dataset, EyePACS, is a U.S.-based teleretinal services provider that provides screening for diabetic eye disease to over 300 clinics worldwide. EyePACS images were acquired as a part of routine clinical care for DR screening, and approximately 40% of the images were acquired with pupil dilation. A variety of cameras were used, including Centervue DRS, Optovue iCam, Canon CR1/DGi/CR2, and Topcon NW using 45-degree fields of view. A subset of the EyePACS clinics recorded HbA1c at each visit. All images and data were de-identified according to HIPAA Safe Harbor prior to transfer to study investigators. Ethics review and Institutional Review Board exemption was obtained using Quorum Review IRB (Seattle, WA).

Retinal fundus images from EyePACS dataset collected between 2007 and 2015 were used as for our development set. For the clinical validation set (EyePACS-2K), we used a random sample of macula-centered images taken at EyePACS screening sites between May 2015 and October 2015 with HbA1c measurements (Table 1). There was no overlap in patients between the EyePACS development set and EyePACS-2K validation set.

**Model development**

A deep neural network model is a sequence of mathematical operations applied to input, such as pixel values in an image. There can be millions of parameters (weights) in this mathematical function[25]. Deep learning is the process of learning the right parameter values ("training") such that this function performs a given task. For example, the model can output a prediction of interest from the pixels values in a fundus image. The development dataset is divided into two components: a "train" set and a "tune" set[1]. During the training process, the parameters of the neural network are initially set to random values. Then for each image, the prediction given by the model is compared to the known label from the training set and parameters of the model are then modified slightly to decrease the error on that image (stochastic gradient descent). This process is repeated for every image in the training set until the model 'learns' how to accurately compute the label from the pixel intensities of the image for all images in the training set. With appropriate tuning and sufficient data, the result is a model general enough to predict the labels (e.g., cardiovascular risk factors) on new images. In this study, we use the Inception-v3 neural network architecture proposed by Szegedy et al[26] to predict the labels.

---

[1] The "tune" set is also commonly called the "validation" set, but to avoid confusion with a clinical validation set (which consists of data the the model did not train on), we are calling it the "tune" set.

We preprocessed the images for training and validation, and trained the neural network following the same procedure as in Gulshan et al[20], but for multiple predictions simultaneously: age, gender, smoking status, BMI, systolic and diastolic blood pressure, and HbA1c. To keep the loss functions on consistent scales, we trained two separate models, one for predicting the binary risk factors (gender, smoking status) and one for the continuous risk factors (age, BMI, blood pressures, HbA1c).

Because the network in this study had a large number of parameters (22 million), we used early stopping criteria[27] to help avoid overfitting: terminate training when the model performance (such as AUC, see statistical analysis section) on a "tuning dataset" stopped improving. The tuning dataset was a random subset of the development dataset that was not used to train the model parameters, but was used as a small evaluation dataset for tuning the model. This tuning set comprised 10% of the UK Biobank dataset, and 2.1% of the EyePACS dataset. To further improve results, we averaged the results of 10 neural network models that were trained on the same data (ensembling[28]).

TensorFlow[29], an open-source software library for Machine Intelligence, was used in the training and evaluation of the models.

**Evaluating the algorithm**

To evaluate the model performance for continuous predictions (age, systolic and diastolic blood pressure, HbA1c), we used the mean absolute error. For binary classification (gender, smoking status), we used the area under the receiver operating curve (AUC). For multiclass classification, we used a simple Cohen's kappa.

**Statistical Analysis**

To assess the statistical significance of these results, we used the non-parametric bootstrap procedure: from the validation set of $N$ patients, sample $N$ patients with replacement and evaluate the model on this sample. By repeating this sampling and evaluation 2,000 times, we obtain a distribution of the performance metric (e.g. AUC), and report the 2.5 and 97.5 percentiles as 95% confidence intervals.

**Mapping Models' Attention**

To better understand how the neural network models arrived at the predictions, we used a deep learning technique called soft attention[30–32] a different neural network model with fewer parameters compared to Inception-v3. These small models are less powerful than Inception-v3, and were used only for generating attention heatmaps and not for the best performances results observed with Inception-v3. For each prediction shown in Figure 2, a separate model with identical architecture was trained. The models were trained on the same training data as the Inception-v3 network described above, and the same early stopping criteria were used.

**Acknowledgements**

From Google Research: Christof Angermueller, PhD, Arunachalam Narayanaswamy, PhD, Alvin Rajkomar, MD, Ankur Taly, PhD, and Philip Nelson.

From EyePACS: Jorge Cuadros, OD, PhD

This research has been conducted using the UK Biobank Resource under Application Number 17643.

**Tables & Figures**

|  | **Development Set** | | **Clinical Validation Set** | |
| --- | --- | --- | --- | --- |
| **Characteristics** | **UK Biobank** | **EyePACS** | **UK Biobank** | **EyePACS-2K** |
| Number of Patients | 48,101 | 236,234 | 12,026 | 999 |
| Number of Images | 96,082 | 1,682,938 | 24,008 | 1,958 |
| Age: Mean, years (SD) | 56.8 (8.2) | 53.6 (11.6) | 56.9 (8.2) | 54.9 (10.9) |
| Gender (% male) | 44.9 | 39.2 | 44.9 | 39.2 |
| Ethnicity | 1.2% Black, 3.4% Asian/PI, 90.6% White, 4.1% Other | 4.9% Black, 5.5% Asian/PI, 7.7% White, 58.1% Hispanic, 1.2% Native Am, 1.7% Other | 1.3% Black, 3.6% Asian/PI, 90.1% White, 4.2% Other | 6.4% Black, 5.7% Asian/PI, 11.3% White, 57.2% Hispanic, 0.7% Native Am, 2% Other |
| BMI: Mean (SD) | 27.31 (4.78) | n/a | 27.37 (4.79) | n/a |
| Systolic BP: Mean, mmHg (SD) | 136.82 (18.41) | n/a | 136.89 (18.3) | n/a |
| Diastolic BP: Mean, mmHg (SD) | 81.78 (10.08) | n/a | 81.76 (9.87) | n/a |
| HbA1c: Mean, % (SD) | n/a | 8.23 (2.14) | n/a | 8.2 (2.13) |
| Current Smoker: % | 9.53% | n/a | 9.87% | n/a |

**Table 1**. Baseline characteristics of patients in the development and validation sets

|  | UK Biobank Validation Set (n=12,026 patients) | | EyePACS-2K Validation Set (n=999 patients) | |
| --- | --- | --- | --- | --- |
| **Predicted Risk Factor (Evaluation Metric)** | Algorithm (95% CI) | Baseline | Algorithm (95% CI) | Baseline |
| Age (MAE in years) | 3.26 (3.22-3.31) | 7.06 (6.98-7.13) | 3.42 (3.23-3.61) | 8.48 (8.07-8.90) |
| Age ($R^2$) | 0.74 (0.73-0.75) | 0.00 | 0.82 (0.79-0.84) | 0.00 |
| Gender (AUC) | 0.97 (0.966-0.971) | 0.50 | 0.97 (0.96-0.98) | 0.50 |
| Current Smoker (AUC) | 0.71 (0.70-0.73) | 0.50 | n/a | n/a |
| HbA1c (MAE in %) | n/a | n/a | 1.39 (1.29-1.50) | 1.67 (1.58-1.77) |
| HbA1c ($R^2$) | n/a | n/a | 0.09 (0.03-0.16) | 0.00 |
| Systolic BP (MAE in mmHg) | 11.35 (11.18-11.51) | 14.57 (14.38-14.77) | n/a | n/a |
| Systolic BP ($R^2$) | 0.36 (0.35-0.37) | 0.00 | n/a | n/a |
| Diastolic BP (MAE in mmHg) | 6.42 (6.33-6.52) | 7.83 (7.73-7.94) | n/a | n/a |
| Diastolic BP ($R^2$) | 0.32 (0.30-0.33) | 0.00 | n/a | n/a |
| BMI (MAE) | 3.29 (3.24-3.34) | 3.62 (3.57-3.68) | n/a | n/a |
| BMI ($R^2$) | 0.13 (0.11-0.14) | 0.00 | n/a | n/a |

**Table 2**. Algorithm performance on predicting cardiovascular risk factors in the two validation sets. 95% confidence intervals on the metrics were calculated with 2000 bootstrap samples (Methods). MAE: Mean Absolute Error; $R^2$: R-squared, AUC: Area under the Receiver Operator Curve (c-statistic). For continuous risk factors (like age), the baseline value is the Mean Absolute Error of predicting the mean value for all patients.

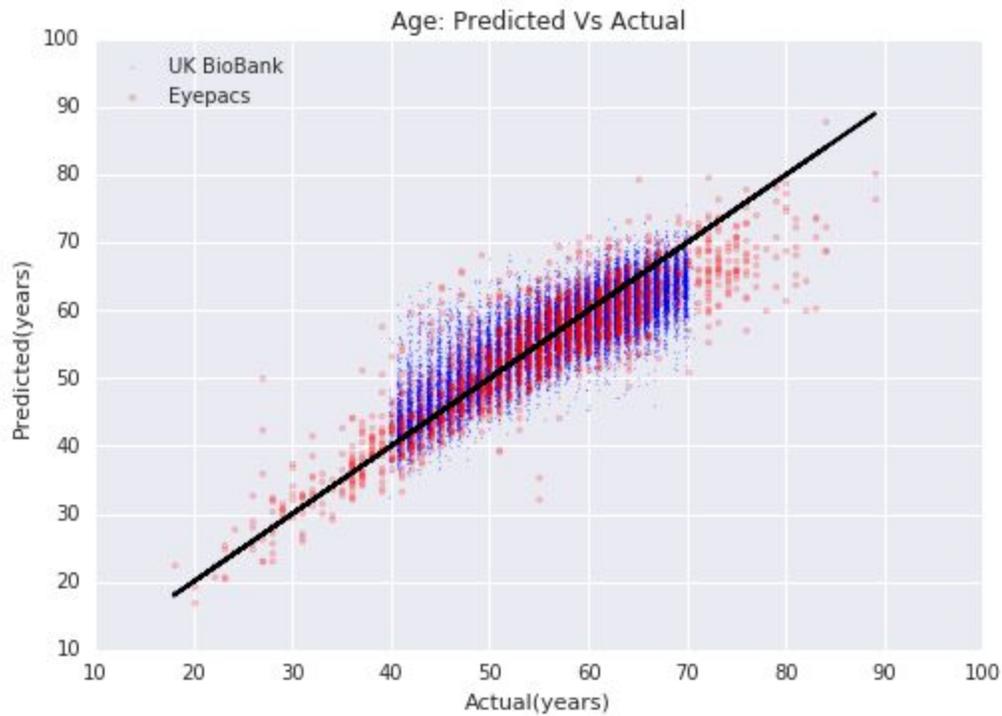

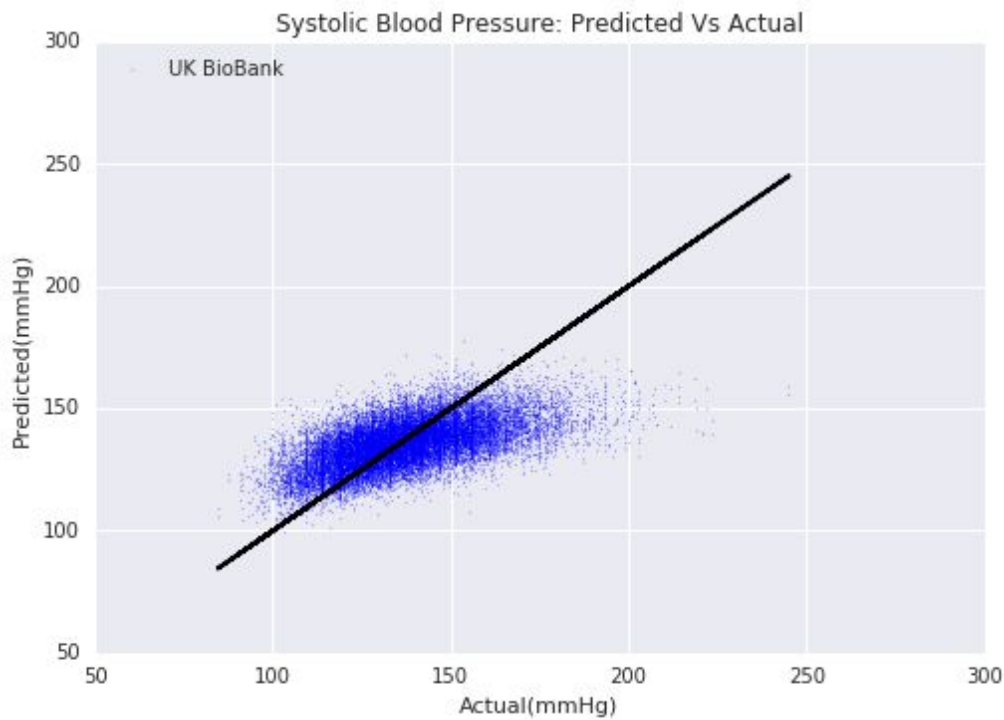

**Figure 1**. (A) Comparing predicted and actual age in the two validation sets, with the y=x line in black. In the UK Biobank validation dataset, age was calculated using the birth year because birth months and days were not available. In the EyePACS-2K dataset, age is available only in units of whole years. (B)

Predicted vs actual systolic blood pressure on the UK Biobank validation dataset, with the y=x line in black.

| Model | AUC (95% CI) |
| --- | --- |
| Age | 0.66 (0.61-0.71) |
| Systolic blood pressure (SBP) | 0.66 (0.61-0.71) |
| Body mass index (BMI) | 0.62 (0.56-0.67) |
| Gender | 0.57 (0.53-0.62) |
| Current smoker | 0.55 (0.52-0.59) |
| Algorithm | 0.70 (0.65-0.74) |
| Age + SBP + BMI + gender + current smoker | 0.72 (0.68-0.76) |
| Algorithm + age + SBP + BMI + gender + current smoker | 0.73 (0.69-0.77) |
| Systematic COronary Risk Evaluation (SCORE)[6,7] | 0.72 (0.67-0.76) |
| Algorithm + SCORE | 0.72 (0.67-0.76) |

**Table 3.** Predicting 5-year MACE on biobank validation set. Of the 12,026 patients in the UK Biobank validation dataset, 91 experience a previous cardiac event prior to retinal imaging and were excluded from the analysis. Of the 11,835 patients in the validation set without a previous cardiac event, 105 patients experienced a MACE within 5 years of retinal imaging. 95% confidence intervals were calculated using 2000 bootstrap samples.

| Original | Age |
|---|---|
| 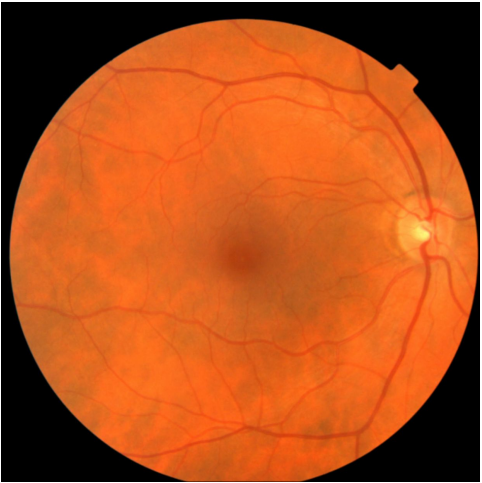 | 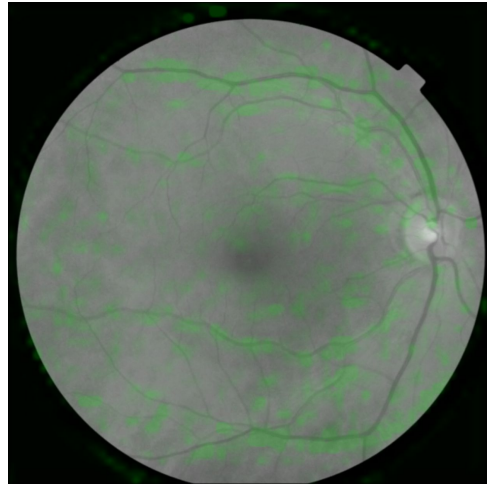<br>Actual: 57.6 years<br>Predicted: 59.1 years |
| **Gender** | **Current smoker** |
| 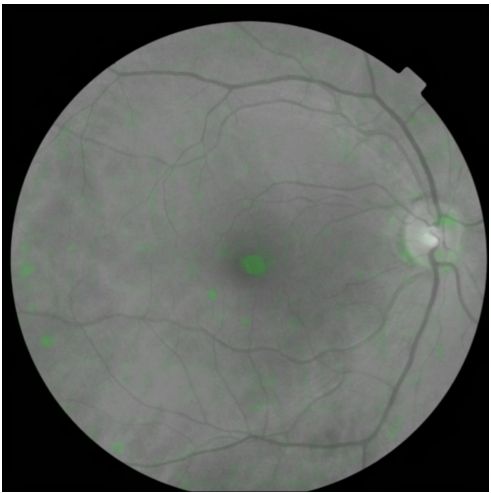<br>Actual: Female<br>Predicted: Female | 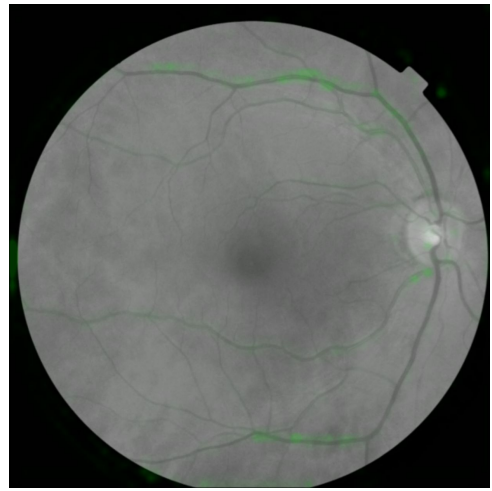<br>Actual: Nonsmoker<br>Predicted: Nonsmoker |
| **HbA1c** | **BMI** |

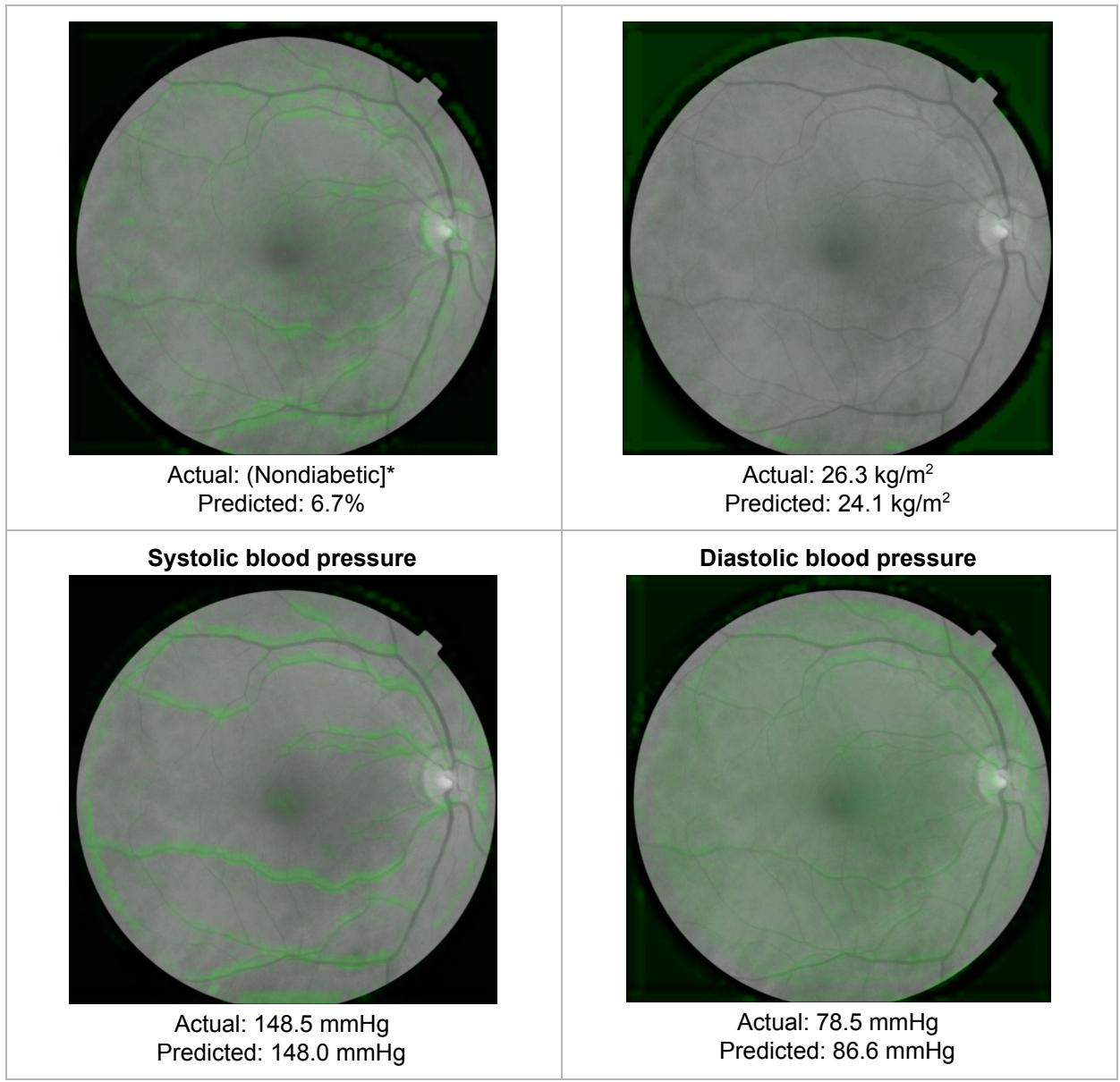

**Figure 2**. **Top left**: Sample retinal image from the UK Biobank dataset in color. **Remaining images**: same retinal image, but in black-and-white. The soft attention heatmap (Methods) for the each prediction is overlaid in green, indicating the areas of the heatmap that the neural network model is using to make the prediction for this image. For a quantitative analysis of what was highlighted, see Table 4. *HbA1c values are not available for UK Biobank patients, so self-reported diabetes status is shown instead.

| Risk factor | Vessels | Optic disc | Nonspecific features |
|---|---|---|---|
| Age | **95%** | 33% | 38% |
| Gender | 71% | **78%** | 50% |
| Current smoker | **91%** | 25% | 38% |
| HbA1c | **78%** | 32% | 46% |
| Systolic BP | **98%** | 14% | 54% |
| Diastolic BP | 29% | 5% | **97%** |
| BMI | 1% | 6% | **99%** |

**Table 4.** Percentage of 100 attention heatmaps where doctors agreed that the heatmap highlighted the given feature. Percentages higher than 75% are bolded. 100 heatmaps were generated for each risk factor, then presented to 3 ophthalmologists who were asked to check off the features highlighted in each image. The images were shuffled and presented as a set of 800, and the ophthalmologists were not told what the heatmaps were intended to explain.